[2024]

NOTICE OF COPYRIGHT

THIS DRAFT OF "**A LOW-COST, HIGH-SPEED, AND ROBUST BIN PICKING SYSTEM FOR FACTORY AUTOMATION ENABLED BY A NON-STOP, MULTI-VIEW, AND ACTIVE VISION SCHEME**" HAS BEEN ACCEPTED BY THE IEEE/RSJ INTERNATIONAL CONFERENCE ON INTELLIGENT ROBOTS AND SYSTEMS (**IROS 2024**). PLEASE CITE THE IEEE VERSION INSTEAD OF THE ARXIV ONE IN YOUR PUBLICATIONS.

# A Low-Cost, High-Speed, and Robust Bin Picking System for Factory Automation Enabled by a Non-stop, Multi-View, and Active Vision Scheme

Xingdou Fu, Lin Miao, Yasuhiro Ohnishi, Yuki Hasegawa, Masaki Suwa

*Abstract*—Bin picking systems in factory automation usually face robustness issues caused by sparse and noisy 3D data of metallic objects. Utilizing multiple views, especially with a one-shot 3D sensor and "sensor on hand" configuration is getting more popularity due to its effectiveness, flexibility, and low cost. While moving the 3D sensor to acquire multiple views for 3D fusion, joint optimization, or active vision suffers from low-speed issues. That is because sensing is taken as a decoupled module from motion tasks and is not intentionally designed for a bin picking system. To address the problems, we designed a bin picking system, which tightly couples a multi-view, active vision scheme with motion tasks in a "sensor on hand" configuration. It not only speeds up the system by parallelizing the high-speed sensing scheme to the robot place action but also decides the next sensing path to maintain the continuity of the whole picking process. Unlike others focusing only on sensing evaluation, we also evaluated our design by picking experiments on 5 different types of objects without human intervention. Our experiments show the whole sensing scheme can be finished within 1.682 seconds (maximum) on CPU and the average picking complete rate is over 97.75%. Due to the parallelization with robot motion, the sensing scheme accounts for only 0.635 seconds in takt time on average.

## I. INTRODUCTION

Bin-picking has been under study both in academia and industry for a long history due to the huge demand for factory automation. Typical and mainstream bin picking systems in factory automation usually utilize 3D sensors as input devices because massive target objects come without texture and most vision controllers are not equipped with GPUs for deep learning on RGB images. Unfortunately, piled metallic objects with a specular surface may cause severe reflection and inter-reflection issues to 3D sensors, which leads to very sparse, noisy, and incomplete 3D data, thus failing 6D pose estimation and the whole picking system afterward. Nevertheless, lots of recent work for picking tries to solve the problem by focusing on 6D pose estimation [1][2][3] instead of input data. Several recent papers [4][5][6-8] are showing the remaining challenges. Since the pose estimation algorithms using 3D data are heavily affected by input data, also stated in [7], we do believe that improving sensing quality is more effective for a robust bin picking system.

Towards this direction, compared with the single-view approach (usually with a multi-shot, phase-shifting 3D sensor

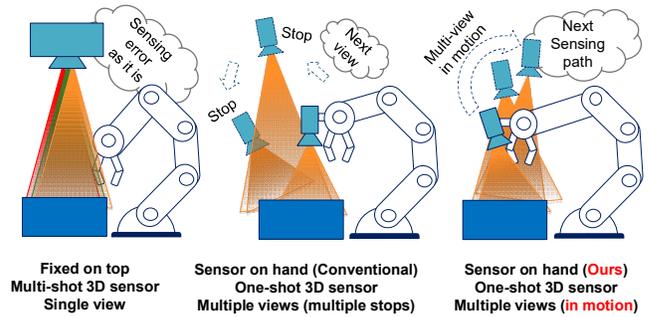

Fig. 1. Typical bin picking systems versus ours. Left: the system comes with an expensive multi-shot 3D sensor fixed on top, leaving the sensing error as it is due to no observations from other views; Middle: the system comes with a cheaper one-shot 3D sensor on robot arm, enabling multi-view sensings (optimization) with multiple stops when approaching objects. Right: our system comes with the same sensor as in the middle, but sensing multiple views (optimization) with robot in motion when leaving objects. Unlike the system in the middle, our sensing is parallel to robot place action and plan continuous sensing path for the next cycle.

[9] fixed on top), multi-view approaches (especially with a one-shot, active stereo camera [10] in a "sensor on hand" configuration) are getting more popularity due to its effectiveness [12-19], flexibility and low cost. Despite the advantages, the multi-view "sensor on hand" bin picking systems [12-19] are still suffering from low-speed issues. That explains why this kind of system is usually seen in labs. In this paper, we try to address this problem, so that low-cost (with a one-shot 3D sensor), multi-view bin picking systems can also be deployed for time-critical applications in factory automation where takt time is highly appreciated.

As is shown in Fig. 1, except for the cost-down from the multi-shot 3D sensor to the single-shot 3D sensor, the "sensor on hand" systems can also capture images from flexible views with robot motion. Furthermore, they can also deal with multiple bins without requiring multiple fixed 3D sensors. With this flexibility, "sensor on hand" systems can enhance sensing quality through volume-based 3D fusion [11-12] or conduct joint optimization [13-17] with multiple views, while fixed systems have no observation from another view to correct errors. As a tradeoff, conventional "sensor on hand" systems move and stop robots, to acquire multiple views passively [11-12] or from predefined locations [13] with redundancy. More complicated ones utilize active vision/sensing path planning to decide the "next best view" based on complex models [8] or virtual rendering [18] in an online manner. These facts show conventional systems just naively apply multi-view sensing to picking systems as a decoupled component from robot motion, ignoring the takt time requirement from various aspects: <1>All these systems follow the serial task order of sensing (approaching target objects), processing, and grasping. It means sensing and processing costs extra time in a picking cycle; <2>The processing such as conventional 3D fusion algorithms [11] involves time-consuming processes such as sensor pose tracking, and volume-based 3D data refining; <3> For perfect hand-eye synchronization, the robot stops to acquire robot kinematics when sensing from each view. The purpose is to transfer 3D data between sensor and robot coordinate systems for robot picking; <4>To decide the "next best view" costs extra processing time, which even deteriorates the waiting time among robot stops; <5> These active vision strategies are designed to improve sensing quality for grasping in the current cycle. They do not guide the robot to observe a suitable area for the next picking cycle, where recognizable target objects may exist. Therefore, the picking process may be discontinued, which is another reason for the low speed. Besides, these sensing paths do not cater to short motion paths as a picking system. To summarize, the mentioned problems are because existing works do not make sensing and motion fully benefit each other as a tightly coupled system. The sensing is not intentionally designed for bin picking systems, whose purpose is to empty the bin at a high speed, instead of perfect sensing results.

As shown in Fig.1 and Fig. 2, targeting a multi-view bin picking system with a one-shot active stereo camera and a "sensor on hand" configuration, we solve these issues with a "tightly coupled" solution enabled by an active vision scheme. The solution includes the following components:

- Firstly, unlike any existing multi-view, "sensor on hand" systems, we reorganize the workflow and parallelize the whole sensing scheme to robot motion (part of place action in "pick and place") without any stops. Our assumption is if the sensing scheme can be finished within the process of place action without any extra cost, the sensing scheme accounts for zero in takt time. Therefore, only reorganizing and parallelization are not enough.

- Secondly, to remove the necessity of multiple stops in multi-view sensing, we designed a unique and high-speed 3D fusion mechanism that can operate efficiently on CPU. We enable continuous multi-view sensing under imperfect hand-eye synchronization (robot in motion) without affecting accurate 3D fusion results in the robot coordinate system. The customized 3D fusion algorithm further speeds up by utilizing robot kinematics for sensor pose tracking as [12] and conducting lightweight data refining on a single depth image instead of a volume-based space.

- Thirdly, to avoid the discontinuity of the picking process, we apply a unique and lightweight active vision strategy to dynamically decide the next sensing path (instead of independent "next best view" at each stop) for continuous sensing. It guarantees the moving camera faces the central area where target objects potentially exist. Our path planning not only improves sensing quality, guarantees the continuity of the whole picking process but also balances the short motion path.

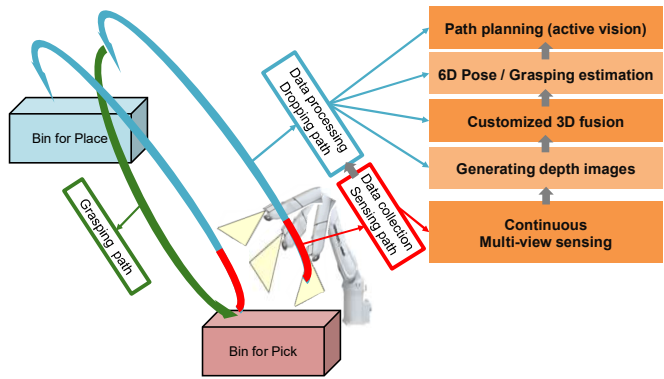

Fig. 2. The workflow and components of our system. The sensing scheme is divided into 2 phases: data collection phase, with robot following sensing path (red); data processing phase, with robot following dorpping path (blue). The system starts with leaving the bin instead of approaching as conventional ones. Parallel to this place action, the system conducts contionus multi-view sensing, 3D data generation from different views, customized 3D fusion, 6D pose/grasping estimiation and path planning. Path planning is for grasping(green), next sensing(red) and next dropping(blue). Our core components are in dark orange, and those in light orange are based on existing approaches.

To our best knowledge, this is the first try to build a high-speed, active vision bin picking system for factory automation. The systematic solution differs from any existing ones and is considered as our biggest contribution in this paper. The sensing components in both data collection and processing are also our contributions, except for the sensor pose tracking [12]. Without loss of generality, it can be customized and applied to a wide scope of pose estimation algorithms and multi-view systems for a dramatic speed-up. Our experiments show the whole sensing scheme which only requires 3 or 4 views can be finished within 1.682 seconds (maximum) on CPU and the average picking complete rate is over 97.75%. Due to the parallelization with robot motion, the sensing scheme accounts for only 0.635 seconds in takt time on average, including the extra motion distance (extra cost) for active sensing.

## II. RELATED WORK

### A. 3D fusion via multiple views

The 3D fusion technique is mainly for SLAM (Simultaneous Localization and Mapping) with a hand-held depth camera. A popular 3D fusion work is from [11], which utilizes vision features to track camera pose in frames and keep a volumetric representation to store and optimize multi-view 3D data. [12] borrows this idea for a picking system. It acquires camera pose directly via robot kinematics for fast speed and higher robustness, but it requires multiple stops for simple hand-eye synchronization. [7] focuses on improving volumetric integration in a probabilistic way by estimating uncertainty maps in a bin picking scenario. Unfortunately, it assumes multiple views come with known camera poses.

### B. Joint optimization via multiple views

[13] applies object segmentation on 2D images captured from 15-18 predefined views, then filters the segmentation masks and projects them to 3D space. The fused 3D points are aligned to pre-scanned 3D models for estimating the 6D pose. [14] makes pose proposals from every single view and keeps those accumulated proposals for future joint

optimization by collision-based pose refinement. [15] estimates the poses of different objects from every single view and then match these object candidates across views by "two-view candidate pair selection". [16] extracts image features from multiple RGB images and performs a 3D point cloud fusion at the same time, then feeds to pose estimation. [17] acquires rotation measurements from different viewpoints and optimizes the rotation estimation with a mixture distribution. These approaches usually require multiple views and require multiple stops. Besides, they lack a policy to guide the sensor to the next view.

*C. Active vision / Sensing path planning*

In addition to accumulating pose proposals and clustering hypotheses from different views for joint optimization, [18] and [19] also use the active vision approach to decide entropy-based "next best view". The purpose is to maximize the sensing information from "selected" views to deal with occlusion and clutter in bin picking. [18] and [19] differ in how to predict the "next best view". [18] is based on virtual rendering and [19] is based on the pose-leaf mapping table built with random forest. [8] digs more into information loss related to surface normal, photometric response, and reflective model of reflective objects. In addition to the demand for multiple stops, these approaches further deteriorate the takt time among robot stops due to complicated active vision policies. Besides, these policies target sensing quality instead of picking continuity.

III. OUR SOLUTIONS

*A. General idea*

As is shown in Fig. 2, under the scope of multi-view, "sensor on hand" picking systems, our solution is based on the unique idea of re-organizing the workflow and parallelizing the whole sensing scheme to robot motion (part of the place action in "pick and place") without stops. It also requires 3 core components:

1) High-speed 3D fusion mechanism, which involves continuous multi-view sensing and customized 3D fusion.
2) Active vision/sensing path planning.
3) Parameter (e.g. view number, robot speed etc.) determination for active vision.

The first is to remove the necessity of multiple stops in multi-view sensing even under imperfect hand-eye synchronization (robot in motion) and speed up 3D fusion processing. The second is to maintain the continuity of the whole picking process.

The whole sensing scheme, with a one-shot active stereo camera, is tightly coupled with robot motion. As is shown in Fig. 2, it is further divided into two phases and related tasks in each. Unlike conventional systems, our picking cycle starts with the robot arm leaving the picking bin instead of approaching it.

1) Data collection phase (including continuous multi-view sensing etc.), with the robot following the sensing path required by active vision.
2) Data processing phase (including 3D generation from stereo images [20], customized 3D fusion, 6D pose/grasping estimation [21], active vision/sensing path planning, etc.), with robot following the dropping path.

We will first introduce how the whole system works with these components and the logic behind the design. Then, we will explain our 3 core components in detail. Components based on existing algorithms [20][21] are skipped.

In data collection phase: the robot arm moves by following the sensing path (the former part of the place action) calculated from the last cycle (the path can be predefined at the first cycle); The sensor on hand captures $n$ pairs of 2D images at time interval $t$. When each pair of images is captured, robot kinematics is also acquired by calling robot SDK, without perfect hand-eye synchronization (due to robot motion). The $n$ pairs of images are not processed to depth images or 3D points in this phase.

One of the purposes of this design is to guarantee the system finishes data collecting within the shortest time and the shortest sensing path. Then, the dropping path (the latter part of place action) can purely cater to the shortest motion path. The sensing path also balances with the short motion which will be explained later. Another purpose is to guarantee the system observes target objects from very close views. This is based on our observations and assumptions but also proved by our experiments. 3D error data of metallic objects are mainly because of direct light reflections or inter-reflections of illumination, so slightly different views see dramatic changes in error data (directions of light reflections). This indicates that 3D fusion with close views is effective in removing errors and filling "missing holes" with a voting mechanism. It is also a reason why our multi-view sensing can speed up without "seeing" around that much.

In data processing phase: the robot arm keeps moving by following the dropping path (the latter part of place action) calculated from the last cycle; In this phase, the system finishes all processing without observation requirement to robot motion, so the dropping path is purely based on the shortest path. Firstly, $n$ depth images from different views are generated from the $n$ pairs of 2D images. Secondly, a customized 3D fusion algorithm fuses these depth images from different views into a single depth image as the target view (e.g. the first view). Thirdly, the 3D fusion result is further fed into a lightweight 6D pose estimation algorithm [21] for fast speed, which optionally takes a depth image as input and only requires CPU. The template-based 6D pose estimation algorithm [21] is suitable since only the piled objects on top are prioritized for recognition in each cycle. After 6D pose estimation, the grasping points are immediately determined since we register the grasping points on CAD models in advance. Fourthly, based on recognized objects in 6D pose estimation and sensor-robot coordinate system transformation, the system decides the grasping path (for the current cycle), sensing path (for the next cycle), and dropping path (for the next cycle) in the robot coordinate system. Among these paths, the sensing path is considered as our active vision strategy, while the grasping path and dropping path are simply based on the shortest path.

*B. High-speed 3D fusion mechanism*

There are 3 bottlenecks for achieving a high-speed 3D fusion in robot vision: <1> sensor pose tracking; <2> sensing with robot stops due to imperfect hand-eye synchronization; <3> volume-based data refining. We will discuss each of them and its corresponding counterplan. Except the first counterplan, the second and the third are all based on our own ideas.

*1) Sensor poses tracking with robot kinematics*

Sensor pose tracking in conventional 3D fusion is based on vision features [11]. It is computationally heavy and may fail in a textureless environment. This can be solved by utilizing robot kinematics, just as in [12]. The sensor pose can be directly acquired through robot pose (end effector pose by calling robot SDK) and hand-eye calibration result.

*2) Continuous sensing without robot stops*

Unlike other systems, we remove the necessity for multiple stops when the sensor captures multiple images along with robot motion. We realize that 3D fusion and sensor-robot coordinate transformation are two different things, which were mixed by other existing works [12][13][14].

3D fusion is to get the "shape" of an object in a reference coordinate system, which corresponds to a specific sensor pose. Comparative rotation and translation between different camera poses are allowed to have slight errors (e.g. due to imperfect hand-eye synchronization) because fusion usually applies ICP (Iterative Closest Point) [22] for refinement at last. Sensor-robot coordinate transformation is to convert the fused "shape" from the reference coordinate system to the robot coordinate system. If the first sensor pose is acquired when the sensor is just about to move (perfect hand-eye synchronization), we can take this camera coordinate system (e.g. the first sensor pose) as the referenced coordinate system. It is used for fused "shape" and as the bridge for further sensor-robot coordinate system transformation. Fig. 3

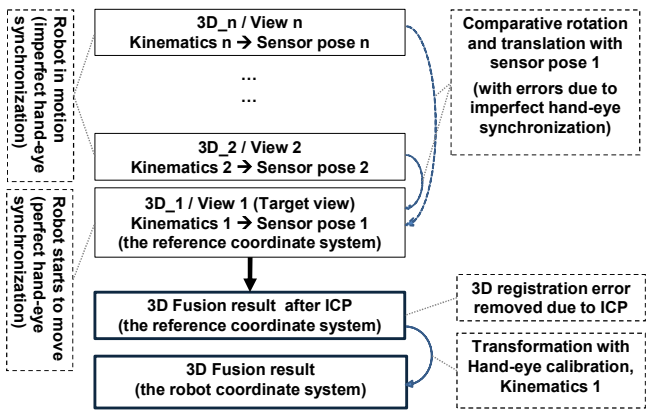

Fig. 3. 3D fusion/Transformation with imperfect hand-eye synchronization data. The system captures *n* views (paired images for depth images, 3D data) and *n* robot kinematics with robot in motion. Depth images from view *n* to view 2 are fused to depth 1 of view 1 based on inaccurate sensor poses, but followed by ICP. Sensor pose 1/view 1 where the perfect hand-eye synchronization exists, is also used as the reference coordinate system to transform the 3D fusion result from sensor to robot coordinate system.

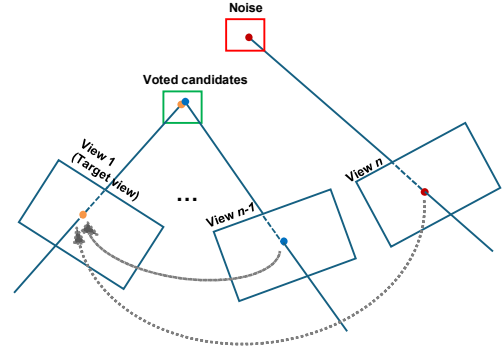

Fig. 4. Our customized 3d fusion on a single depth image for fast speed. Depth images from view 2…view *n* with their corresponding 3D point sets and senor poses are reprojected to view 1 as target view. The pixel coordinate in target view can be used as the index receiving candidates and voting. The 3D points in orange and blue are in close distance (less than 2*sensor accuracy) and their averaged value is considered as voted result.

explains the method in detail.

In most picking systems, after the gripper grasps an object in the bin, there is an action to "pull up" the object out of the bin for obstacle clearance and then quickly place it in another location. At the end of the "pull-up" action, the robot stops very shortly. We set this moment to capture the first view in our multi-view capturing process, and the corresponding sensor pose is considered as the reference coordinate system.

*3) 3D Fusion on a single depth image*

3D fusion with conventional volumetric representation, such as TSDF (Truncated Signed Distance function) [11] or octree-based approach [14], requires huge memory and causes heavy computational costs to store and refine data. Since we do not need to keep all data for mapping or rendering, we fuse all depth images from different views into a single depth image of the target view. As is seen in Fig. 4, pixels in each depth image can be converted to 3D points and then projected to the target view. Each pixel coordinate in the target depth image serves as an index to receive multiple depth candidates. Finally, each pixel can be refined with voting, as shown in a pseudo program of algorithm 1.

| **Algorithm 1**: To fuse depths from other views into a single depth of the target view |
|---|
| **Input**: depth_1… n, sensor_pose_1…n<br>**Output**: depth_11 |
| 1. [depth_11…depth_1n] = Project [depth_1…depth_n, sensor_pose_1…sensor_pose_n, ICP] |
| 2. For each pixel (x,y) in each [depth_11…depth_1n] |
| 3.    distance = Compare (pixel(x,y), candidates(x,y)) |
| 4.    If distance < threshold *δ (2*sensor accuracy)* |
| 5.       Average (candidate(x,y), pixel(x,y)) |
| 6.       Increase_vote(candidates(x,y)) |
| 7.    Else |
| 8.       Insert (pixel(x,y), candidates(x,y)) |
| 9. For each (x,y) in depth_11 |
| 10.  depth_11(x,y) = FindMaxVotes (candidates(x,y)) |
| 11. Return (depth_11) |

## C. Active vision / Sensing path planning

### 1) The definition of sensing path

As shown in Fig. 5, we need to determine two sensor poses, the starting pose and the ending pose, for a sensing path. The sensor always faces the target objects and captures multiple images in between. For either starting or ending pose, we need to determine two properties:

- Target center (Where to look at, "star" in Fig. 5 represents its center). The two poses share the same center of the target area. The interpolated path between these two poses (a common function inside robot SDK) helps to guarantee that the sensor at different poses in between still faces the target area.
- Location (From where to look at the target). The distance between the location and the target center should be within the sensor working distance.

Angle β is a parameter called "coverage angle". It guarantees the robot to finish capturing the expected number of images by going through the sensing path.

### 2) Determine a sensing path for the next cycle

To decide the target center, we use an "object-oriented" strategy. An example is given in Fig. 6, the left image shows the sensing and recognition result in the current cycle. There are four objects recognized in green and orange and the one in green is selected to be grasped. The target center is set to the weighted gravity center of the recognized objects which will remain in the next cycle, and calculated by:

$$\text{Target center} = \frac{\sum_{k=1}^{n}[Confidence_k * X_k]}{\sum_{k=1}^{n} Confidence_k} \quad (1)$$

where $n$ is the number of recognized objects (except the one to be grasped), $X_k$, and $Confidence_k$ are the position and 6D pose estimation confidence of the object $k$ respectively. The percentage, representing the confidence in each recognized object, is directly from the pose estimation approach [21]. With this active vision, the system dynamically shifts its sensing orientation to the weighted area where objects with proven recognition success in the last cycle may potentially exist. It effectively maintains the continuity of the whole picking process as shown by our experiments.

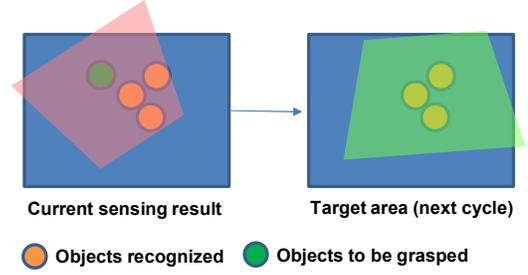

Fig. 6. The method to decide target center for sensing path in next cycle. Left: the current sensing and recognition results (the object in green is to be grasped, the objects in orange will remain.) Right: the weighted gravity center of the remaining objects in orange is set to be the target center. Our active vision strategy guides the sensor to observe the area where objects are likely recognizable and potentially exist for next sensing.

To decide the sensor locations, we search for two points $M$ (starting pose location) and $N$ (ending pose location) on the intersection between a 3D sphere $S$ (Guarantee all points in the best sensor working distance to the target area) and a 3D plane $P$ (Limit the sensing path inside it to shorten the motion path), as shown in Fig. 7. $M$ and $N$ with the target center $T$ comprise the coverage angle β.

- 3D sphere $S$: Centered at point $T$, the center of the target area; the radius is the best working distance of the 3D sensor.
- 3D plane $P$: The target object is to be picked from point $A$ and to be lifted out of the box vertically to point $B$. Point $B$ is vertically above the bin at a certain fixed distance (e.g., 10cm). After that, the robot moves from $B$ to $C$ (the dropping point), going through the sensing path MN and the dropping path NC. $A$, $B$, and $C$ decide the 3D plane $P$.

Angle γ is a predefined constant parameter (e.g. 5 degrees) to help the sensing path get tilted to the dropping point, to shorten the motion path. The pseudo program of algorithm 2 shows the process of finding the locations of the two poses.

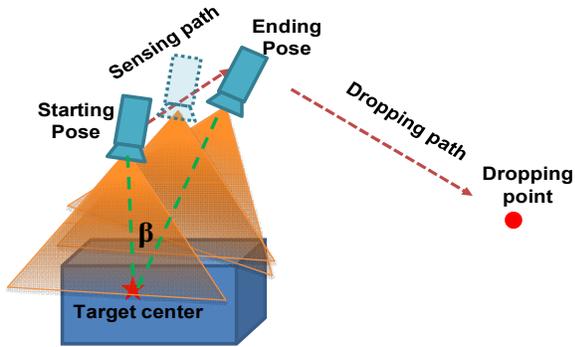

Fig. 5. The definition of our sensing path. It includes two poses, starting pose and ending pose. Each pose includes its target center and location (within sensor working distance). The two poses share the same target center so that interpolated poses in between along with the sensing path also face the target center ("star"). Angel β is a parameter to make sure the expected number of views are captured within the sensing path.

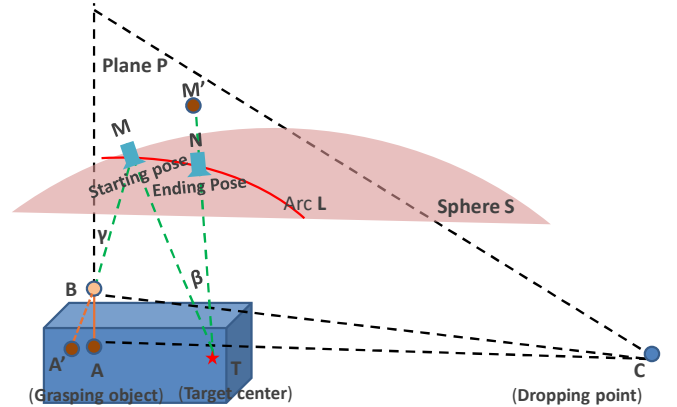

Fig 7. The geometry and the method of determining locations of starting pose and ending pose. Two locaions $M$ and $N$ are searched on the intersection (Arc $L$) of the sphere $S$ (centered at target center with the radius of best sensor working distance) and plane $P$ (decided by grasping object $A$, its lift-up $B$ and dropping location $C$). The "coverage angel" β is to make sure the expected number of views are captured in between. Angle γ is set to 5 degrees as a constant.

| **Algorithm 2**: To find the locations of starting pose and ending poses for a sensing path |
|---|
| **Input**: 3D sphere S, 3D plane P, Point A, B, <br> **Output**: Point M, N |
| 1. Arc L = Intersect (S, P) <br> 2. Line A'B = Rotate (Line AB, by B, angle γ, *within P*) <br> 3. Point M = Intersect (A'B, L) <br> 4. Line M'T = Rotate (Line MT, by T, angle β, *within P*) <br> 5. Point N = Intersect (M'T, L) <br> 6. Return M, N |

*D. Active vision parameters (e.g., view number…)*

   *1) Definition of sensing path parameters*

As seen in the right of Fig. 8, within the sensing path we assume that the robot moves at a constant speed $v$ and $n$ images are captured at time interval $t$. A more natural way to understand it is from the left of Fig. 8, the arc (sensing path) degree angle β and the interval angle α that averagely divides this arc into $n$-1 parts. Angle β is called the "coverage angle". The two sets of parameters are interchangeable. We assume the same type of randomly piled objects exhibit similarities on the subsurface, so the same parameters are applied.

   *2) Determining sensing path parameters*

The parameters are determined through an offline process and are set only once for each type of object. It takes less than 15 minutes since we can remove redundant data collection and process data in parallel. Motivated by a high-speed sensing scheme, we aim to find the optimal parameters that achieve the minimum view number, the fastest robot moving speed, and the highest number (ideal) of recognized objects. We first determine robot speed $v$, time interval $t$, and the number of views $n$ with the experiments, then change them to angle α and β for further use. Due to symmetry, we move the sensor randomly on a 1/8 sphere surface (center: a certain location of piled objects, radius: sensor best working distance) and keep it facing the sphere center. During the movement, the sensor keeps capturing images (50 in total) with different $v$ and $t$ (discrete values: v<=80% every 10%, t<=80ms every 10ms). Actually, $t$ is fixed to 10*ms* since its data can propagate to 20…80 *ms*. This process is repeated 10 times with random paths. The data will be processed with different $n$ (limited to *10*) images/views till 6D pose estimation. We take the number of recognized objects as a criterion for evaluation. The steps to decide parameters are as follows: <1>Assume the numbers of recognized objects follow a Gaussian distribution. Filter the data with the range from σ to 2σ to avoid noise. <2>Filter the rest of the data with the most frequent value of $n$. <3> Find the highest value of $v$ from the rest. Speed is represented as a percentage in our test robot. We roughly proportionate 100% to 1 meter/second.

## IV. EXPERIMENTS

*A. Hardware configuration*

In our experiment, as you can see in Fig. 9, we attach a one-shot active stereo camera (view area: 75cm*50cm @ best working distance 30cm), similar as Intel Realsense D435 or Ensenso N30 (single-shot mode), to an OMRON Adept robot arm S650. The projector of the 3D sensor is customized to keep illumination always on when the 3D sensor continuously captures multiple images in a short time. Due to strong active illumination, the exposure time for all target objects is set to 3 ms to avoid blur. The sensor is connected to a PC (Intel® Core™ i5-7440EQ, 8 GB DDR4) without GPU for all vision processing, which simulates a typical vision controller. A program running on a robot motion controller can acquire robot kinematics and set an I/O to trigger our 3D sensor for capturing images (built-in memory). For picking, a two-finger gripper and a vacuum gripper are used for different types of target objects.

*B. Target objects*

Fig. 10 shows the target objects with sizes. They all have high reflectance rates but differ in shape to cover diversity. For each type of object, we pick 20 objects within a box whose size is around 50cm*25cm.

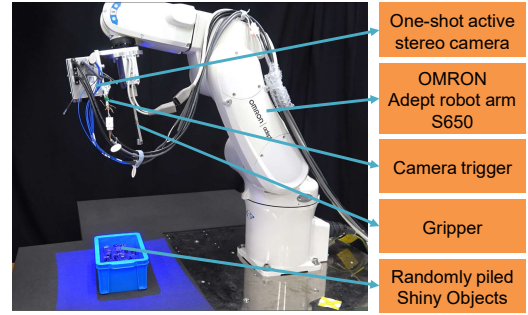

Fig. 9. The hardware of our bin picking system. The picking system comes with a customized one-shot active stereo camera (view area 75cm*50cm @ 30cm) on the top of a robot arm (OMRON Adept S650). Two-finger gripper and vacuum gripper are used for different types of shiny objects randomly piled in a box (50cm*25cm).

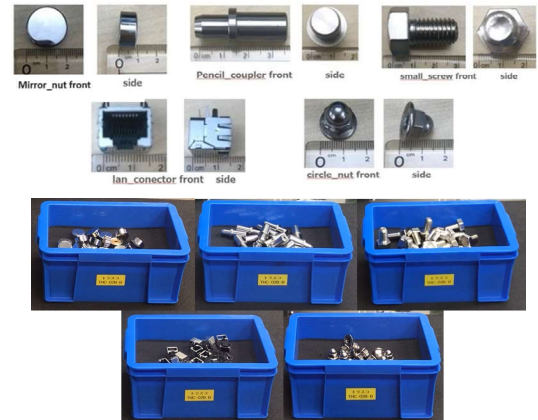

Fig 10 . Top: the 5 different target objects with shiny surfaces and different shapes. Bottom: 20 piled target objects in each box (50cm*25cm) for picking experiments.

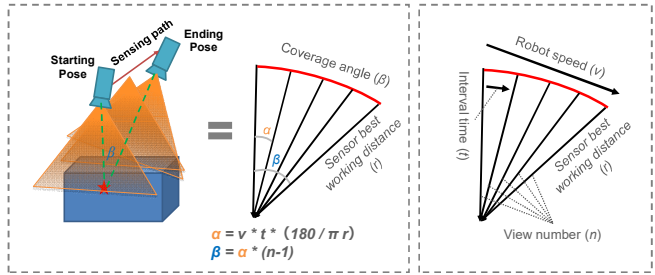

Fig. 8. The description of sensing path parameters. Two ways (left and right) to define parameters, but interchangeable. Left: the parameters are defined as angle α and β. β is called coverage angel to guarantee sensor capture $n$ views with interval α along the sensing path. Right: the robot moves at a constant speed $v$ and $n$ images are captured at time interval $t$.

## C. Time of the whole sensing scheme

Table I shows the way to calculate the time for the whole sensing scheme. *n* is the number of views and *t* is the time interval for image capture. The total time can be calculated by:

$$\text{Total time} = 343n+t*(n-1) + 220 \text{ ms} \quad (2)$$

where the max in our picking test (*n*=4, *t*=30) is 1682 ms.

TABLE I. TIME ESTIMATION OF THE WHOLE SENSING SCHEME

| *Paired 2D images* | *Kinematics* | *3D* | *Estimation* | *Path planning* |
|---|---|---|---|---|
| 1. Exposure<br>2. Transfer<br>3. Interval | Aquirement and transfer | 1. Each view<br>2. 3D fusion | 6D pose and grasp | Sensing, dropping, and grasping |
| $3n+90n+t*(n-1)$ | $20n$ | $150n+80n$ | 200 | 20 |
| **Total**: $343n+t*(n-1) + 220$  /  **Total max** (*n*=4, *t*=30): = 1682 ms<br>(*n*: view number, *t*: time interval, unit: milliseconds) ||||| 

## D. Experiments

We focus on three indicators: Sensing results (between two baselines A, B, and ours); Sensing time in takt time (between typical "sensor on hand" systems and ours); Picking complete rate of our picking test (5-type objects).

*1) A (Single view + active vision) versus ours (multi-view + active vision)*

As in Fig. 11, this experiment (target objects: small screw) shows that the sensing and recognition result (in green) using a single view (in the left collum) compared with that using multiple views (in the right column, 4 views). With a smaller number of objects, the single view sensing is more prone to totally failed recognition, which may discontinue the picking process frequently. In this case, if 7 out of 20 objects are left, the picking complete rate is 65%.

*2) B (Random sensing + multi-view) versus ours (active vision + multi-view)*

As in Fig. 12, this experiment (target objects: the first row, mirror nut / the second row, small screws) shows that the sensing and recognition result (in green) using random sensing (in the left collum) compared with that using active vision (in the right column). Our active vision always makes the sensor face the central area where objects with proven recognition success in the last cycle exist, for better sensing results. For the case in the first row, where 9 out of 20 objects are left, the picking complete rate is only 45%.

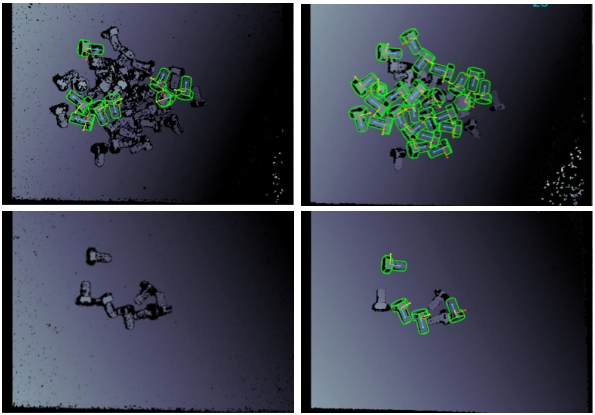

Fig. 11. The result of baseline A (single view + active vision) versus ours. Objects in green represent successful 6D pose estimation. Left column: result of A; Right column: ours. Top row (small screw): more objects case; Bottom row (small screw): less objects case. Ours always gets better recognition results due to multi-view sensing.

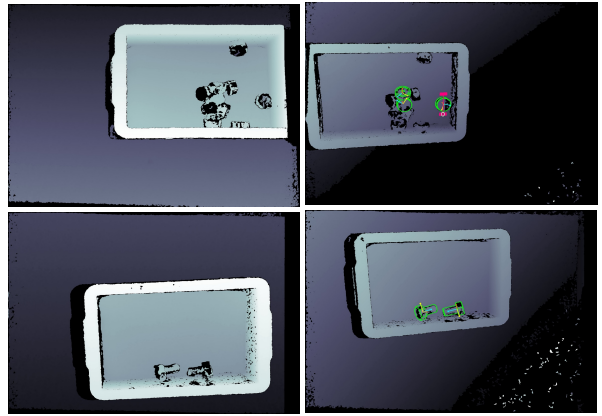

Fig. 12. The result of baseline B (random sensing + multi-view) versus ours. Objects in green represent successful 6D pose estimation. Left column: result of B; Right column: ours. Top row (mirro_nut) and Bottom row (small screw) are two different types of objects. Ours always observes the objects with proven recognition success in the last cycle and from the view center for better sensing results.

*3) Sensing time cost in takt time (Conventional "sensor on hand" systems versus ours)*

Table II shows the estimated sensing time of several multi-view "sensor on hand" picking systems [6-8][12][14][16] from their papers or videos. Unfortunately, [6-8][16] assume multiple views with known camera poses already exist even most of them target picking applications. If the sensing is for multiple objects pickings, averaged time is used. None of them parallelize the sensing and processing with robot motion as ours. Our sensing scheme is faster than place action (assuming average place action > 2 seconds [12][14]), so it can be finished in parallel. It accounts for only 0.635 (instead of 1.682) seconds in takt time on average in our picking tests, including the extra motion distance for active sensing (Difference between trajectory B-M-N-C and B-C in Fig.7). Even sensing costs 1 second in takt time, (assuming grasp and place costs 4 seconds.), 10 hours can lead to 1800 picking cycles difference.

TABLE II. SENSING TIME COST IN TAKT TIME

| **Robot Fusion [12]** | **More Fusion [14]** | **MV6D, ROBI-related [16], [6-8]** | ***Ours*** |
|---|---|---|---|
| ≈6.0 s | ≈8.0 s | Not applicable | ≈0.635* s |
| * Due to the parallelization with robot motion, our sensing scheme accounts for only 0.635 seconds in takt time on average, including the extra motion distance for active sensing. ||||

*4) Picking test*

Picking test with single view or random sensing is skipped since either of them frequently fails the recognition and the picking process, especially when half objects remain. We conduct 20 times picking tests for each type of target object directly with our approach. In each experiment, the robot proceeds picking experiments on 20 objects without any human intervention. As is shown in Table III, target objects only require 3~4 views. In 100 (20*5) experiments, all objects (20*5*20) are successfully recognized till the last one. Searching is rarely triggered (averagely < 1 time for the whole picking process) to recover the failed recognition. Failed picking actions are due to the interference between the gripper and the box. The average picking complete rates for 5 target objects are 100%, 97.75%, 98.75%, 100%, and 100%. With carefully designed gripers and grasping policy (not our focus in this paper), the picking complete rate can be further improved. Fig. 13 shows an example (target object: Lan connector) of sequential status in different picking cycles within the same experiment.

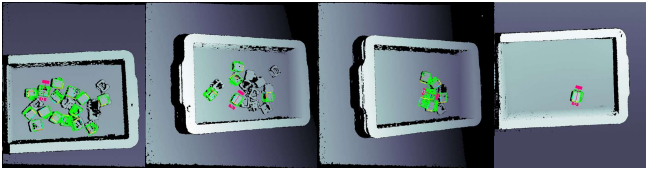

Fig. 13. An example (target object: lan connector) of sequential status in different picking cycles within one experiment. From left to right: the 1st, 5th, 15th and last cycle.

TABLE III.  PICKING TEST RESULTS

| Gripper | Two-finger | | | | Vacuum |
|---|---|---|---|---|---|
| Objects | Small screw | Lan connector | Mirror nut | Pencil coupler | Circle nut |
| Sensing Parameters (view number, interval, speed) | 4 views 30ms 70% | 4 views 30ms 80% | 3 views 60ms 60% | 4 views 30ms 80% | 4 views 30ms 80% |
| Average complete rate | 100% | 97.75% | 98.75% | 100% | 100% |
| Lowest complete rate | 100% | 90% | 90% | 100% | 100% |
| Highest complete rate | 100% | 100% | 100% | 100% | 100% |

## V. Conclusion

In this paper, to fill the high-speed gap between existing multi-view "sensor on hand" picking systems and time-critical industry picking applications, we designed a unique sensing scheme, which is tightly coupled with robot motion, to form a fast active vision picking system as the first try. There is still space to improve the sensing scheme to approach 1 second. Without loss of generality, our idea can be customized and applied to a wide scope of pose estimation algorithms and multi-view systems for a dramatic speed-up. In the future, we want to adapt the system to other multi-view approaches (e.g. those with multiple 2D images instead of 3D fusion) for generalization.